%% file: main.tex
\documentclass{article} 
\usepackage{iclr2024_conference,times}

\input{math_commands.tex}

\usepackage{hyperref}
\usepackage{url}
\usepackage{graphicx}
\usepackage{cleveref}
\usepackage{booktabs} 
\usepackage{multirow}
\usepackage{marvosym}
\title{Hierarchical Auto-Organizing System for Open-Ended Multi-Agent Navigation}

\author{Zhonghan Zhao$^{*,1}$, Kewei Chen$^{*,2}$, Dongxu Guo$^{*,2}$, Wenhao Chai$^{\dagger,3}$, Tian Ye$^{*,4}$, \\
\textbf{Yanting Zhang$^{2}$, and Gaoang Wang$^{1,\text{\Letter}}$}\\
$^{1}$ Zhejiang University \quad
$^{2}$ Donghua University \quad
$^{3}$ University of Washington\\
$^{4}$ Hong Kong University of Science and Technology (GZ) \\
\texttt{\{zhonghan.22, gaoangwang\}@intl.zju.edu.cn, wchai@uw.edu}
}

\iclrfinalcopy 
\begin{document}

\maketitle
\renewcommand{\thefootnote}{\fnsymbol{footnote}}
\footnotetext[1]{Equal contribution, \textsuperscript{$\dagger$}Equal engineering contribution, \textsuperscript{\Letter} Corresponding author. 
\newline This work is supported by the National Natural Science Foundation of China (No. 62106219, No. 62206046), the Zhejiang Provincial Natural Science Foundation of China (No. LZ24F030005), and the Shanghai Sailing Program (No. 21YF1401300).}
\renewcommand*{\thefootnote}{\arabic{footnote}}

\begin{figure*}[h]
\centering
    \includegraphics[width=\linewidth]{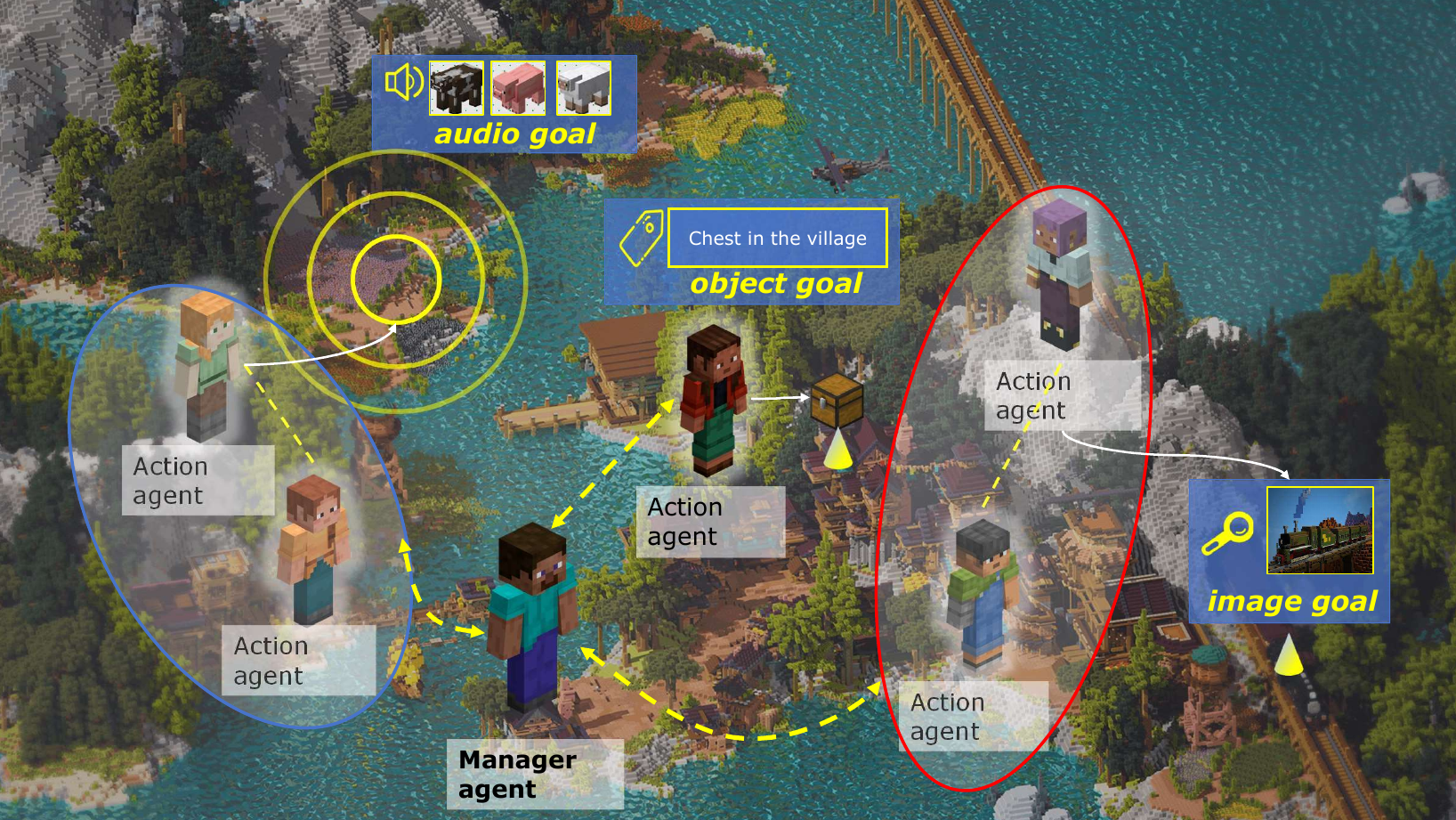}\\
    \caption{Illustration of the functionality of \textbf{HAS}, in the Minecraft environment.
    Given a navigation goal in the form of images, objects, or audio, collectives of agents autonomously self-organize to undertake collaborative endeavors of navigation like searching and exploring.
    }
    \label{fig:fig_teaser}
\end{figure*}

\input{sections/0_abstract}
\input{sections/1_intro}
\input{sections/2_relatedwork}
\input{sections/3_method}
\input{sections/4_experiments}

\input{sections/5_conclusion}

\bibliography{ref}
\bibliographystyle{iclr2024_conference}
\end{document}

%% file: math_commands.tex
\usepackage{amsmath,amsfonts,bm}

\def\eqref#1{equation~\ref{#1}}

\def\1{\bm{1}}

\DeclareMathAlphabet{\mathsfit}{\encodingdefault}{\sfdefault}{m}{sl}
\SetMathAlphabet{\mathsfit}{bold}{\encodingdefault}{\sfdefault}{bx}{n}



%% file: sections/0_abstract.tex
\begin{abstract}
Due to the dynamic and unpredictable open-world setting, navigating complex environments in Minecraft poses significant challenges for multi-agent systems. Agents must interact with the environment and coordinate their actions with other agents to achieve common objectives. However, traditional approaches often struggle to efficiently manage inter-agent communication and task distribution, crucial for effective multi-agent navigation. Furthermore, processing and integrating multi-modal information (such as visual, textual, and auditory data) is essential for agents to comprehend their goals and navigate the environment successfully and fully. To address this issue, we design the HAS framework to auto-organize groups of LLM-based agents to complete navigation tasks. In our approach, we devise a hierarchical auto-organizing navigation system, which is characterized by 1) a hierarchical system for multi-agent organization, ensuring centralized planning and decentralized execution; 2) an auto-organizing and intra-communication mechanism, enabling dynamic group adjustment under subtasks; 3) a multi-modal information platform, facilitating multi-modal perception to perform the three navigation tasks with one system. To assess organizational behavior, we design a series of navigation tasks in the Minecraft environment, which includes searching and exploring. We aim to develop embodied organizations that push the boundaries of embodied AI, moving it towards a more human-like organizational structure.
\end{abstract}

%% file: sections/1_intro.tex
\section{Introduction}
Drawing on large language models (LLMs) with human knowledge, communicative competence~\cite{mao2020learning,mao2020learning}, and decision-making capabilities, embodied agents exhibit human-like intelligence in playing games~\citep{park2023generative, wang2023voyager, zhao2023see}, programming~\citep{qian2023experiential, hong2023metagpt}, and robotic tasks~\citep{zhang2023building, mandi2023roco}. 
The development of individual intelligence has led to creating a new, collaborative framework where multiple agents~\cite{chen2023autoagents, chen2023agentverse} work together. 

This shift towards a multi-agent system~\cite{hong2023metagpt, chen2023autoagents} uses advanced language understanding and decision-making abilities to enhance intelligence through interactions and data sharing. Agents in this system specialize in various tasks, sharing insights to enhance overall efficiency. This collaborative approach not only optimizes execution but also fosters a learning environment where agents continually improve through shared intelligence. Their interaction leads to refined skills, equipping them to handle complex, large-scale tasks through parallel processing and coordinated collaboration, a notable advancement in autonomous systems.

Multi-modal navigation~\citep{wijmans2019dd,yu2021sound,du2020vtnet,kwon2023renderable,chen2021history,moudgil2021soat} stands at the forefront of contemporary AI research, representing a rapidly evolving field that aims to integrate various sensory inputs into a cohesive navigational strategy. This emerging domain extends the capabilities of multi-agent systems by requiring them to interpret and act upon a confluence of sensory data. 
Unlike traditional methods focusing on rendered images or static virtual environments, works on dynamic environments~\cite{deng2023citygen}, such as multi-modal navigation in open-ended environments like Minecraft, pose a more complex challenge. Agents in Minecraft~\cite{wang2023voyager, zhao2023see} navigate a world full of freedom and variability, making it an ideal testbed for embodied agent systems. Minecraft's unpredictable and open-ended environment is an ideal testing ground for advanced AI systems. These systems emulate human adaptability and intelligence, processing and sharing multi-modal data to strategically navigate and complete tasks. This approach goes beyond simple improvements, marking a significant leap in developing autonomous systems that handle complex tasks with remarkable autonomy and skill.
Integrating Multi-modal Language Models (MLMs) into embodied agents enhances navigational efficiency. Equipped with LLMs, these agents engage intelligently with their environment. The zero-shot performance capabilities of LLMs enable them to interpret and act upon data without explicit programming. This makes them ideal for open-ended environments.

Our vision is centered on employing MLM-based embodied agents to revolutionize navigation within the intricate landscapes of Minecraft. Environments in this game present unique challenges, such as the need to respond to image, audio, and object cues in real-time, as highlighted in~\Cref{fig:fig_teaser}. By harnessing the zero-shot learning capabilities and the nuanced decision-making abilities of LLMs, our agents can navigate these spaces with efficiency and adaptability akin to human intuition. Such versatility is achieved without requiring exhaustive retraining or complex reconfigurations, marking a substantial leap towards autonomous systems that engage with their surroundings in more sophisticated and human-like ways.

We summarize our contributions as follows:

\noindent \textbf{$\bullet$} 
We introduce \textbf{HAS}, a hierarchical structure for multi-agent navigation based on LLMs in the Minecraft environment. It utilizes centralized planning with decentralized execution, enabling efficient multi-modal navigation in open-ended environments. 

\noindent \textbf{$\bullet$} 
We design an auto-organizing and intra-communication mechanism to dynamically adjust the key role and action group based on the task allocation and maintain inter-group communication to ensure efficient collaboration.

\noindent \textbf{$\bullet$} 
We achieve state-of-the-art performance on the asynchronous multi-modal navigation task on image, audio, and object goals in Minecraft's open-ended environment. 

%% file: sections/2_relatedwork.tex
\section{Related Works}

\subsection{Intelligent Agent in Minecraft}
As an open-ended sandbox game, Minecraft has always been an ideal setting for testing the performance of intelligent agents~\citep{johnson2016malmo,hofmann2019minecraft}. The agents must autonomously perform various tasks in Minecraft, such as chopping trees, crafting tools, and mining diamonds. At the beginning, much of the works focus on exploring reinforcement learning~\citep{lin2021juewu,mao2022seihai,skrynnik2021hierarchical,lifshitz2023steve} or imitation learning~\citep{amiranashvili2020scaling,baker2022video}, without satisfactory performance. VPT~\citep{baker2022video} and MineDojo~\citep{fan2022minedojo} collect internet-scale datasets for their model pre-training. VPT enables direct learning to act during video pre-training and using these behaviors as exploration priors for reinforcement learning.
Yet, recent works found that the pre-trained LLMs could serve as a strong ``mind'' that provides planning ability to the agents. Voyager~\citep{wang2023voyager} is a single-robot multiple agent system 
that leverages multiple groups of GPT-4~\citep{openai2023gpt4} as a high-level planner, low-level action code generator, critic generator, and curriculum manager. Plan4MC~\citep{yuan2023plan4mc} proposes a skill graph pre-generated by the LLMs. DEPS~\citep{wang2023describe}, an interactive planning method based on LLMs,  addresses multi-step reasoning issues in open-world planning. GITM~\citep{zhu2023ghost} develops a set of structured actions and leverages LLMs to generate action plans for the agents to execute,  achieving impressive results in various tasks. 

\subsection{Embodied Multimodal Model}

Embodied agents integrate sensory perceptions, physical actions, and computational intelligence to accomplish tasks and goals within their environment. Key areas are wide-ranging, including Navigation~\citep{wijmans2019dd,yu2021sound,du2020vtnet,kwon2023renderable,chen2021history,moudgil2021soat}, Embodied Question Answering~\citep{das2018embodied,yu2019multi,datta2022episodic}, Active Visual Tracking~\citep{luo2018end,zhong2021towards,luo2019end,zhong2019ad}, and Visual Exploration~\citep{liu2022symmetry,dean2020see,chen2018learning}. The field is evolving rapidly with the development of Large Language Models (LLMs)~\citep{song2022llm} and Multimodal LLMs (MLLMs)~\citep{alayrac2022flamingo,zhu2023minigpt,li2023otter,li2022blip,li2023blip,gong2023multimodal,lyu2023macaw,ye2023mplug,dai2023instructblip,wang2023visionllm,liu2023visual,maaz2023video,su2023pandagpt,gao2023llama}, integrating multiple modalities for more effective processing. A prime example of this innovation is PaLM-E~\citep{driess2023palm}, a sophisticated multimodal model with 562B parameters, adept at a broad spectrum of embodied tasks and demonstrating exceptional capabilities in visual reasoning.

\subsection{LLM-based Multi-Agent Frameworks} 

Large Language Models~(LLMs) are skilled at completing new tasks when given prompt-based instructions. Autonomous agents based on Large Language Model-based~(LLM-based) models have gained significant interest in industry and academia~\citep{wang2023survey}. 
Several works~\citep{wang2023unleashing,du2023improving,zhuge2023mindstorms,hao2023chatllm,akata2023playing,zhang2023controlling} have augmented the problem-solving abilities of LLMs by incorporating discussions among multiple agents. Stable-Alignment~\citep{liu2023training} generates instruction datasets by reaching a consensus on value judgments through interactions among LLM agents in a sandbox. Some works in the field of artificial intelligence focus on studying sociological phenomena. For instance, Generative Agents~\citep{park2023generative} creates a virtual ``town'' comprising 25 agents to investigate language interaction, social understanding, and collective memory. The Natural Language-Based Society of Mind~\citep{zhuge2023mindstorms} involves agents with different functions interacting to solve complex tasks through multiple rounds of Mindstorms. In addition, others~\citep{cai2023large} propose a model for cost reduction by combining large models as tool makers and small models as tool users.

Some works emphasize cooperation and competition related to planning and strategy~\citep{meta2022human}, some propose LLM-based economies~\citep{zhuge2023mindstorms}, and others propose LLM-based programming~\citep{hong2023metagpt}. Developing an LLM-based multiple embodied agent system with control capability is challenging. We face difficulties in multi-agent cooperation, such as low efficiency of global perception and generalizing dynamic groups. To overcome these challenges, we aim to apply a hierarchical auto-organizing system in multi-agent navigation.

%% file: sections/3_method.tex
\section{HAS: A Multi-agent Navigation Framework}
\begin{figure*}[h]
\centering
    \includegraphics[width=\linewidth]{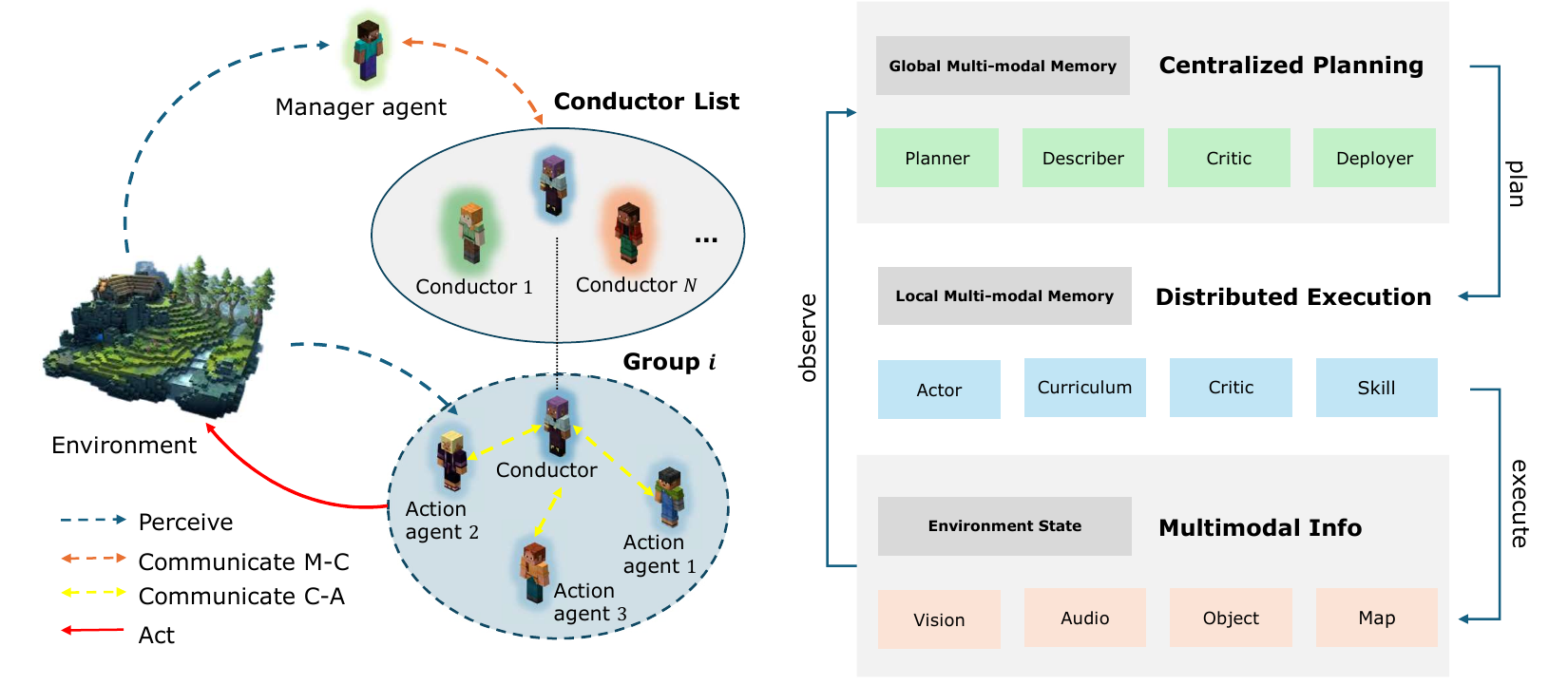}\\
    \caption{\small \textbf{HAS framework.} M-MLM and A-MLM correspond to the memory-augmented multi-modal language models of the manager and action agent, respectively. It also utilizes a multi-modal memory to store and obtain experiences as references for planning. HAS can improve its planning skills by exploring its own proposed tasks with self-instruction and using its growing memory to plan tasks it has visited before.
    }
    \label{fig:framework}
\end{figure*}

As shown in~\Cref{fig:framework}, the \textbf{HAS} is a hierarchical LLM-based multi-modal multi-agent navigation framework denoted as $\mathcal{F}$, which can manage and execute complex multi-agent navigation tasks on image~($Im$), object~($Oj$), and audio~($Au$) goals with perception on the state list of vision~($v$), audio~($a$), and other properties~$p$ within open-ended environments by leveraging cognitive and collaborative capabilities of the multi-modal language model~($\mathbf{MLM}$):
\begin{equation}
\mathbf{a}^c_i = \mathcal{F}(s,g,\mathcal{M}),\\\ s = {(v^c_i, a^c_i, p^c_i)}_{i\leq N}, \\\ g \in G=\{Im, Oj, Au \}
\end{equation}
where $(v^c_i, a^c_i, p^c_i)$ represents the state list of vision, audio, and other properties of one conductor agent, \( g \) is the original goal, and \( \mathcal{M} \) is the dynamic map. Then, we get \( \mathbf{a}^c_i\) as the action for the conductor agent \( i\). 

The Hierarchical architecture consists of two primary operational domains: higher-order centralized planning, which is managed by the manager multi-modal language model ($\mathbf{MLM}^M$), and ground-level decentralized execution, which is conducted by the conductor model ($\mathbf{MLM}^C$), then the action $a^c_i$ of the conductor can be obtained as follows,

\begin{equation}
\mathcal{F}(s,g,\mathcal{M}) = \mathbf{MLM}^C(\mathbf{MLM}^M(s,g,\mathcal{M}), s_i),  \quad s_i = (v^c_i, a^c_i, p^c_i)
\label{eqn:main_function}
\end{equation}

where $\mathbf{MLM}^C$ and $\mathbf{MLM}^M$ represent the multi-modal language models of the conductor and manager agents.

\subsection{Centralized Planning with Decentralized Execution}\label{sec:CPDE}

Evoked from the Centralized Training with Decentralized Execution~(CTDE) framework~\citep{hu2023attention,sunehag2017value,lowe2017multi,mao2018modelling} for cooperative Multi-Agent Reinforcement Learning~(MARL), we propose Centralized Planning with Decentralized Execution~(CPDE) for the cooperative LLM-based Multi-Agent system. This architecture leverages global state information to inform the planning process, while the execution of tasks is carried out by local agents independently. CPDE is tailored to maximize the synergy of global oversight and local autonomy, ensuring efficient navigation and task completion in complex environments.

\paragraph{Centralized planning for manager agent.}

The centralized planning process for a manager agent $M$ simulates a high-level understanding like that of a human planner. The process includes understanding the environment's dynamic global states, recognizing the conductor agents' capabilities and limitations, and devising a strategy. It contains $\mathbf{MLM}^M$ consisting of 4 MLM modules with different functions for the manager as mentioned in~\Cref{sec:MLM} and a Global Multi-modal Memory for storing multi-modal information as mentioned in~\Cref{sec:mmm}.

\paragraph{Decentralized execution for action agent.}

The decentralized execution process is designed to capitalize on the autonomy and flexibility of conductor agents $C$. These agents navigate the environment, perform tasks, and learn from their interactions guided by the strategic direction from the centralized planning of the manager agent. Note that these conductors can deploy several action agents for similar goals of one sub-goal. It contains $\mathbf{MLM}^C$ comprising 4 MLM modules for conductors with different functions as mentioned in~\Cref{sec:MLM} and a Local Multi-modal Memory for storing multi-modal information as mentioned in~\Cref{sec:mmm}.

\paragraph{Auto-organizing mechanism.}

The auto-organizing mechanism is a spontaneous grouping mechanism to promote multi-agents' efficiency. In the centralized planning phase, the manager agent auto-organizes several conductor agents based on the global environment and tasks, which takes advantage of the planning capabilities of MLM and is inspired by AutoAgents~\citep{chen2023autoagents} to deploy agents with different roles automatically. During the decentralized execution stage, we are inspired by the Self-Organized Group (SOG)~\citep{shao2022self} to solve the issue of zero-shot generalization ability with dynamic team composition and varying partial observability. We improve and use a novel auto-organizing mechanism. Each group has conductor-action agent consensus, where the action agents can only communicate with their conductor. Additionally, we utilized the summarization ability~\citep{ma2023large} from LLM to summarize and distribute the messages received to all affiliate group members to hold a unified schedule.
 
\subsection{Multi-modal Language Model} \label{sec:MLM}
The Multi-modal Language Model $\mathbf{MLM} = \{\mathbf{MLM}^M, \mathbf{MLM}^C\}$, consisting of two different types of MLM of the manager and conductor agents. $\mathbf{MLM}^M = \{\mathcal{P}_l, \mathcal{D}_s, \mathcal{C}_r^m, \mathcal{D}_p\}$, which consists of Planner, Describer, Critic, and  Deployer for the manager. They formulate aligned task plans, condense and translate multi-modal data, refine strategies through feedback, and assign and direct agent subtasks respectively.
$\mathbf{MLM}^C = \{\mathcal{A}_c, \mathcal{C}_u, \mathcal{C}_r^c, \mathcal{S}_k\}$, which consists of Actor, Curriculum, Critic, and  Skill module for conductors. They translate strategic plans into executable actions, orchestrate dynamic group formations, and distribute tasks across agents, ensuring alignment with centralized directives and facilitating continuous learning and adaptation through a curriculum of complex tasks.

\paragraph{Adaptive planning with MLM.}

Our approach integrates the environment's observations and task directives to plan actions based on the current scenario. We begin by translating multi-modal observations into textual descriptions, utilizing a method that avoids direct scene captioning by the MLM. Instead, we extract keywords for items from the STEVE-21K dataset~\citep{zhao2023see} and employ GPT-4 to craft sentences that articulate these observations. The MLM identifies relevant condition sentences from textual observations during the planning phase. It also incorporates additional context, such as biome types and inventory levels, into text formats via predefined templates. We generate action plans by re-engaging the MLM's linguistic component with the task instructions and these descriptive texts. This methodology leverages the MLM's capabilities in a layered manner, yielding more accurate situational descriptions and plans that significantly reduce the likelihood of generating unrealistic elements compared to fully integrated models.

\paragraph{Autonomous error correction and proactive planning.}

HAS enhances its planning through a closed-loop feedback mechanism, automatically correcting failures by analyzing feedback and identifying errors using its self-explanation capabilities. Unlike other agents, it generates improved plans without human input or extra information. Additionally, HAS simulates and evaluates each plan step to identify potential flaws early, reducing the likelihood of encountering difficult situations due to plan failures. This proactive approach enables it to foresee issues like insufficient resources, which could hinder task completion.

\subsection{Multi-modal Memory}\label{sec:mmm}
Research~\citep{hong2023metagpt} has shown that memory mechanisms play a crucial role in the functioning of generalist agents. Equipping HAS with multi-modal memory enables it to plan using pre-existing knowledge and real-world experiences, improving planning accuracy and consistency. The MLM in HAS allows leveraging these experiences in context without requiring additional model updates. 

We have illustrated the design of our multi-modal memory system. At a high level, this system is a key-value memory model with multi-modal keys comprising both the task and the observation of the state when this memory entry is created. The values stored in this memory system are the plans that were successfully executed. Since the plans are in an open-ended environment, Minecraft, there could be multiple entries with the same task but different observations and plans. As a result, HAS needs to generate multi-modal queries based on the current task and situations to retrieve the relevant memory entries.

\paragraph{Retrival-augmented storage.}
Retrival-augmented storage~(RAS) enables long-term planning capability by Retrieval-Augmented Generation (RAG)~\citep{lewis2020retrieval,mao2020generation}. RAG improves the quality of responses generated by language models with external sources of knowledge to complement the model's internal representation. Instead of external knowledge libraries, we use the collected multi-modal memory as the knowledge library and retrieve interactive experiences as demonstration prompts to improve the planning results.
The formulation is as follows:
\begin{align}
    p(y \mid x) \approx  \sum_{z\in \text{top-k}(p(\cdot \mid x))}p_\eta (z \mid x) \cdot p_\theta (y \mid x,z),
\end{align}
where $x$, $y$, and $z$ represent instructions, plans, and retrieved memory entries. $p_\eta$ and $p_\theta$ denote retrieval and planning models. This retrieval-augmented planning method helps HAS to ground its internal knowledge into open-ended environments efficiently. It also leverages historical interaction feedback to solve hallucinations within LLMs and produce more accurate plans.

\paragraph{Multi-modal retrieval.}
Multi-modal retrieval~(MMR) enables efficient access to a rich repository of multi-modal memories. This process is initiated with a query containing textual and visual elements. To align this query with the trajectories stored within the multi-modal memory, we utilize the manager's Describer $\mathcal{D}_s$. The Describer converts visual information into a textual format. This textual description serves as an image tag, amalgamating with other textual data or as a textual representation for audio information.

When a retrieval request is made from the multi-modal memory, especially when the information is an amalgamation of image and text, the Describer module $\mathcal{D}_s$ is employed to transcribe the image into text. Subsequently, this description is used to compute the similarity across the multi-modal memory entries. The top-k most similar entries are retrieved for further processing. The formalization of this retrieval process is as follows:
\begin{align}
    \mathcal{R}(q_t, q_v) = \sigma(\mathcal{D}_s(q_v), q_t),
\end{align}
where $q_t$ denotes the textual query, $q_v$ represents the visual query, and $\mathcal{R}$ signifies the retrieval function. The function $\mathcal{S}$ computes the similarity between the multi-modal memory entries and the query, using the textual description provided by $\mathcal{D}_s$.

\paragraph{Dynamic map.}
The dynamic map visually represents the exploration domain, showcasing only the areas the agents have explored. It is the main way for the manager agent to see the global environment. When using the map, it will be used in the form of an image. It is updated in real-time with information from all action agents, providing a strategic overview of the environment. This map is instrumental in planning and executing navigation tasks as it reflects the current knowledge and discoveries made by the agent collective.

The following equations can formalize the dynamic map's function:

\begin{equation}
    \mathcal{M}_t = \mathcal{M}_{t-1} \cup \bigcup_{i=1}^{n}S_{i,t}, \qquad S_{i,t} = F(\mathbf{txt}_{i,t})
\end{equation}

Where \( M_t \) represents the dynamic map at time \( t \), \( S_{i,t} \) is the state information from agent \( i \) at time \( t \), including text data $\mathbf{txt}_{i,t}$ like the place name, special materials. \( F \) is the function that integrates the text data into the map, and \( n \) is the number of active agents. This real-time updating mechanism ensures that the dynamic map remains an accurate and current representation of the exploration field.

%% file: sections/4_experiments.tex
\section{Experiment}
Our experiments aim to achieve three goals using HAS in challenging Minecraft navigation tasks. Firstly, we want to evaluate the performance of HAS against baselines that do not fully address the issues faced by open-world agents. Secondly, we aim to understand the factors that contribute to these results. Lastly, we want to explore the potential of HAS for life-long learning and its benefits in long-horizon tasks. We will first introduce the evaluation settings, present the main comparative results and ablation studies, and conclude with an exploratory trial on long-horizon tasks.

\subsection{Experimental Setups}
We select gpt-4-1106-vision-preview~\citep{yang2023dawn} as the base model. Our simulation environment is based on MineDojo~\citep{fan2022minedojo} and uses Mineflayer~\citep{mineflayer} APIs for motor controls. The maximum number of robots that can be allocated based on this environment is 8, which is also our experimental robots' upper limit.

\subsection{Baselines}
Currently, no MLM-driven multi-agents~(robots) work out of the box for Minecraft, so we carefully selected several representative algorithms as baselines for our experiment. They rely on extracting information from a system's backend, presenting a significant divergence from real-world scenarios. 

\paragraph{Voyager}\citep{wang2023voyager} is a blind, single-robot system, relying only on textual grounding for perception. It has a long-term procedural memory that stores a hierarchical library of code-based grounding procedures. Complex skills can use simpler skills as sub-procedures. Voyager is known for its ability to explore areas and master the tech tree. However, its main focus is to prompt GPT-4~\citep{openai2023gpt4} on background text messages in embodied agents. We use multiple hosts to deploy multiple models to work directly on the server. We convert the input of the image goal into a text task through GPT-4V, using the same Describer module as ours.

\paragraph{STEVE}\citep{zhao2023see} is a multi-modal single-robot system that combines the vision unit with the STEVE-13B with the code database. It focuses on introducing a visual module to endow the model with visual perception capabilities in processing visual perception information and handling task reasoning for skill-related execution. Similarly, using multiple hosts, we deploy multiple models to work directly on the server. This single-agent method optimizes performance. We also convert the input of the image goal into a text task by using the same Describer module as ours.

\subsection{Task setting}
We set the world on peaceful mode and agents always start in survival mode with an empty inventory so that navigation tasks can be performed without interruption. Meanwhile, we choose over 15 environment seeds of 6 different terrains from the STEVE-21K dataset~\citep{zhao2023see} for evaluation. The tasks we chose mainly test the efficiency of long-distance directivity navigation, short-range non-directional navigation, and free-world exploration.

A task is considered successful when the target object is close to the target or when a certain number of targets are reached.  Due to the open-world nature of Minecraft, the world and initial position that the agent is spawned at could vary a lot. Therefore, we conduct at least 30 tests for each task and report the average efficiency and success rate to ensure a thorough assessment. 

\vspace{-10pt}

\subsection{Evaluation Results}

\paragraph{Multi-modal goal search.}
\input{tables/goal_search}
Multi-modal goal search includes goals of Image, Object, and Audio.
Object labels identify in-game items such as villages, pyramids, and animals. Image labels help locate objects using images. Audio labels are used to detect sounds outside of the player's range. Due to an insufficient audio library, we set the perceptible range around the target and passed it to the agent as text. It is similar to finding objects at close range, expressing distance through feedback intensity values. 

As shown in~\Cref{tab:goal_search}, \textbf{HAS} achieves the best performance as a multi-agent system. Due to the auto-organizing mechanism, multiple agents can be better planned than directly adding multiple free agents. Even with a single-agent system, our performance is still the best. This is because our system decomposes tasks layer by layer, allowing both managers and action agents to maintain focus on their specific tasks. With the help of dynamic maps, the agent can be dynamically updated to understand the environment better globally. Our method has sufficient potential for performance growth. With the improvement of robots, it does not require too many robots to find a reasonable number of robots in the audio goal at close range.

\vspace{-10pt}

\paragraph{Continuous block search.}
\input{tables/block_search}
Continuous block search is a close exploration mission to assess the agent's exploratory capabilities and proficiency in locating multiple blocks in a row. Diamond blocks are placed at every 16-block interval across the mainland map.

As shown in~\Cref{table:block_search}, we experiment with the block-searching task~\citep{zhao2023see} to assess the agent's exploratory capabilities and proficiency in locating specified blocks. Dynamic map is to identify as many blocks as possible within the fewest iterations, which indicates the method's efficiency. The dynamic map and the self-organization of the hierarchical structure allow more blocks to be found and deployed.
\vspace{-10pt}

\paragraph{Map exploration.}
\input{tables/map_explore}
Map exploration aims to let the agent update the map as much as possible. We set up the same status awareness: when in an unreached area, status information prompts the agent in text. We set each step's maximum movement distance not to exceed 50 blocks. 

As shown in~\Cref{table:map_explore}, \textbf{HAS} has achieved the best performance, especially on multi-agent, which can better reflect the hierarchical structure. It is of considerable value for macro-control of multi-agent exploration and avoids agents repeatedly wandering in the explored area.

\subsection{Ablation Study}
\input{tables/ablation}

To understand the impact of different components on the performance of our system, we conducted ablation studies. The results, as shown in Table~\ref{tab:ablation}, provide insights into the effectiveness of the dynamic map and auto-organizing mechanism. Note that in w/o AO, we directly block the planner and deployer modules of the Manager to remove the auto-organizing mechanism and only retain the describer's perception of the environment, including dynamic maps into status information, thereby directly acting on the same target task on a fixed number of agents.
\vspace{-10pt}
\noindent\paragraph{Dynamic map is critic.}
Dynamic map is a multi-modal information body including images and status information. It can integrate data in the simplest way to provide global understanding, reduce additional overhead and provide the possibility of macro layout for multi-person cooperation.
\vspace{-10pt}
\noindent\paragraph{Comparison with GPT-4.}
Auto-organizing mechanism can dynamically adjust dynamic groups and automatically assign tasks based on the existing environment, allowing efficient multi-person cooperation.

%% file: tables/goal_search.tex
\begin{table}[!t]
\centering
\resizebox{\linewidth}{!}{
\begin{tabular}{@{}l|c|cc|cc|cc@{}}
\toprule
\multirow{2}{*}{Method} & \multirow{2}{*}{\# agents} & \multicolumn{2}{c|}{\bf Image Goal} & \multicolumn{2}{c|}{\bf Object Goal} & \multicolumn{2}{c}{\bf Audio Goal}\\
\cmidrule(lr){3-4} \cmidrule(lr){5-6} \cmidrule(lr){7-8}
& & \# iters~($\downarrow$) & success rate~($\uparrow$) & \# iters~($\downarrow$) & success rate~($\uparrow$) & \# iters~($\downarrow$) & success rate~($\uparrow$)  \\
\midrule
\multirow{2}{*}{Voyager} & 1 & 95 & 0.21 & 64 & 0.41 & 21 & 0.67 \\
                 & 3 / 2 / 5 & 45 & 0.47 & 36 & 0.59 & 6 & 0.85 \\
\midrule
\multirow{2}{*}{STEVE} & 1 & 85 & 0.25 & 71 & 0.31 & 13 & 0.71 \\
              & 5 / 5 / 4  & 32 & 0.52 & 29 & 0.57 & 6 & 0.82 \\ 
\midrule
\multirow{2}{*}{\bf HAS~(Ours)} & 1 & 27 & 0.76 & 15 & 0.83 &4 & 0.87\\
                        & 8 / 7 / 3 & \textbf{6}  & \textbf{0.84} & \textbf{4}  & \textbf{0.95} &\textbf{2} & \textbf{0.99}\\
\bottomrule
\end{tabular}
}
\caption{\textbf{Comparison on goal search task.} \# iters represent the average number of iterations required to finish each task (find goals) with a maximum of 100 prompting iterations~(only within the statistical range). The success rate is for task fulfillment. We list the one-agent and best performance with the number of agents~(from left to right represents different goals) on the bottom line for each method according to \# agents. Note that due to dynamic robots, this number for our method refers to the peak number.}
\label{tab:goal_search}
\end{table}

%% file: tables/block_search.tex
\begin{table}[t!]
\centering
\resizebox{0.5\linewidth}{!}{
\begin{tabular}[c]{l|ccc}
\toprule
Method & \# agents & \# iters~($\downarrow$) & \# blocks~($\uparrow$) \\ 
\midrule
\multirow{2}{*}{Voyager}\citep{wang2023voyager}  & 1 & 34 & 28\\
                                                 & 3 & 14 & 81\\
\midrule
\multirow{2}{*}{STEVE}\citep{zhao2023see} & 1 & 15 & 65\\
                                          & 4 & 6 & 211\\
\midrule
\multirow{2}{*}{\textbf{HAS}} & 1 & 14 & 68\\
                              & 8 & \textbf{2} & \textbf{367}\\
\bottomrule
\end{tabular}
}
\caption{\textbf{Comparison on continues block search task.} \# iters represent the average number of iterations required to locate the first ten diamond blocks with a maximum of 100 prompting iterations; the lower this number, the higher the task completion efficiency. \# blocks denotes the average number of diamond blocks found over 100 iterations, with higher values indicating better performance. We list the one-agent and best performance with the number of agents on the bottom line for each method according to \# agents.
}
\label{table:block_search}
\end{table}

%% file: tables/map_explore.tex
\begin{table}[t!]
\centering
\resizebox{0.5\linewidth}{!}{
\begin{tabular}[c]{l|ccc}
\toprule
Method & \# agents & \# iters~($\downarrow$) & area~($\uparrow$) \\ 
\midrule
\multirow{2}{*}{Voyager}\citep{wang2023voyager}  & 1  & 6 & 175 \\
                                                  & 5  & 3 & 755 \\
\midrule
\multirow{2}{*}{STEVE}\citep{zhao2023see} & 1  & 6 & 161 \\
                                           & 6  & 3 & 696\\
\midrule
\multirow{2}{*}{\textbf{HAS}} & 1 & 1 & 201 \\
                              & 8 & \textbf{1} & \textbf{1368}  \\
\bottomrule
\end{tabular}
}
\caption{\textbf{Comparison on map exploration task.} \# iters represent the average number of iterations required to locate the area of 100 blocks; the lower this number, the higher the task completion efficiency. \# area denotes the average squares of blocks over 5 iterations, with higher values indicating better performance. We list the one-agent and the best performance with the number of agents on the bottom line for each method according to \# agents.
}
\label{table:map_explore}
\end{table}

%% file: tables/ablation.tex
\begin{table}[!t]
\centering
\resizebox{\linewidth}{!}{
\begin{tabular}{@{}l|c|cc|cc|cc@{}}
\toprule
\multirow{2}{*}{\bf Setting} & \multirow{2}{*}{\# agents} & \multicolumn{2}{c|}{\bf Goal Search} & \multicolumn{2}{c|}{\bf Block Search} & \multicolumn{2}{c}{\bf Map Exploration}\\
\cmidrule(lr){3-4} \cmidrule(lr){5-6} \cmidrule(lr){7-8}
& & \# iters~($\downarrow$) & success rate~($\uparrow$) & \# iters~($\downarrow$) & \# blocks~($\uparrow$) & \# iters~($\downarrow$) & area~($\uparrow$)  \\
\midrule
\multirow{2}{*}{w/o DM} & 1 & 53 & 0.46 & 14 & 67 & 6 & 160 \\
                & 6 / 4 / 5 & 22 & 0.64 & 5 & 237 & 3 & 624\\
\midrule
\multirow{2}{*}{w/o AO} & 1 & 41 & 0.55 & 35 & 29 & 6 & 172 \\
               & 5 / 5 / 5  & 15 & 0.78 & 11 & 106 & 3 & 706 \\ 
\midrule
\multirow{2}{*}{\bf HAS~(Ours)} & 1 & 15 & 0.82 & 14 & 68 & 1 & 201\\
                        & 6 / 8 / 8 & \textbf{4}  & \textbf{0.93} & \textbf{2} & \textbf{367} & \textbf{1} & \textbf{1368}\\
\bottomrule
\end{tabular}
}
\caption{\textbf{Ablation studies} for multi-modal goal search, continuous block search, and map exploration. The setting is the same as the above 3 experiments. Note that w/o DM is without the dynamic map, and w/o AO is without the auto-organizing mechanism.}
\label{tab:ablation}
\end{table}

%% file: sections/5_conclusion.tex
\section{Conclusion}
In conclusion, \textbf{HAS} brings a significant advancement to multi-agent systems for complex environment navigation. Our contributions include a hierarchical auto-organizing system that ensures effective centralized planning and decentralized execution, an innovative auto-organizing and intra-communication mechanism for dynamic task adaptation, and a multi-modal information platform that integrates diverse sensory inputs. These advancements collectively enhance the autonomy, efficiency, and adaptability of AI agents, marking a pivotal step forward in the field of Embodied AI.